# BIDRN: A Method of Bidirectional Recurrent Neural Network for Sentiment Analysis

**Dr. D Muthusankar[1], Dr. P Kaladevi[2,*], Dr. V R Sadasivam[3], R Praveen[4]**

[1,2]Department of Computer Science and Engineering, K. S. Rangasamy College of Technology, Tiruchengode, Tamil Nadu, India
e-mail: muthusankar@ksrct.ac.in; Kaladevi@ksrct.ac.in*

[3]Department of Information Technology, K. S. Rangasamy College of Technology, Tiruchengode, Tamil Nadu, India
e-mail: sadasivam@ksrct.ac.in

[4]Departmentof Computer Technology, Anna University, MIT Campus, Chennai, Tamil Nadu, India.
e-mail: Praveenram2510@gmail.com

**Abstract**—Text mining research has grown in importance in recent years due to the tremendous increase in the volume of unstructured textual data. This has resulted in immense potential as well as obstacles in the sector, which may be efficiently addressed with adequate analytical and study methods. Deep Bidirectional Recurrent Neural Networks are used in this study to analyze sentiment. The method is categorized as sentiment polarity analysis because it may generate a dataset with sentiment labels. This dataset can be used to train and evaluate sentiment analysis models capable of extracting impartial opinions. This paper describes the Sentiment Analysis-Deep Bidirectional Recurrent Neural Networks (SA-BDRNN) Scheme, which seeks to overcome the challenges and maximize the potential of text mining in the context of Big Data. The current study proposes a SA-DBRNN Scheme that attempts to give a systematic framework for sentiment analysis in the context of student input on institution choice. The purpose of this study is to compare the effectiveness of the proposed SA- DBRNN Scheme to existing frameworks to establish a robust deep neural network that might serve as an adequate classification model in the field of sentiment analysis.

**Keywords**- Opinion Mining, Text mining, Sentimental analysis, Jallikattu, Sentiment Polarity, Neural Networks.

## I. INTRODUCTION

Opinion mining referred to as sentiment mining which is a potential domain of research that tries to determine the opinion lying underneath a text represented in natural language. This opinion mining is an integrated discipline of computational linguistics and information retrieval. It aids in estimating and extracting subjective information from the related source materials. It concentrates on the process of mining opinions from the collection of source document that portrays significant information about an object. The subjective opinion mining is achieved by extracting attributes of the objects from the social network user comments to identify whether the extracted opinions are positive, neutral, and negative [1,2,3]. Moreover, the degree of interest exhibited by the individual users towards online opinions related to a social event, political happening, products, and services plays a vital tool for perceiving the mentality of the people towards the respective issue. These opinions are determined to introduce comprehensive impact for the purpose of opinion sharping, such that the government, political parties, financial institutions, public and private companies, and organization can monitor them from effectively perceiving the mind of the common people [4,5,6]. However, the process of web monitoring is a complex activity as there exists a diversified number of sources that contain voluminous data [7,8,9].

At this juncture, Opinion mining tools are considered as a boon for potentially processing a huge number of online reviews for the objective of determining the inherent opinions. Furthermore, the machine learning approaches that extracts and explores the opinions on the domain feature level for reliable classification of overall opinion on a multi-scale is essential [10,11,12]. The deep neural network architectures of machine learning approaches are identified to be potential in extracting and investigating huge number of opinions on the level of the domain attributes with the precise classification of sentiments into different degrees of sentiment polarity. In this research, three different deep learning techniques are proposed and compared to identifying the superior among them in terms of accuracy in classifying sentiments [13,14,15].

Text is analyzed in this work to get human insights and characteristics of social network users [16]. Post polarities are grouped into feelings such as positive feedback, negative feedback, and comparable ones. Sentiment analysis is divided into two categories: lexical analysis and machine learning analysis [17]. Lexical analysis focuses on calculating the degree of polarity of a specific content based on the semantic positioning of the words or the phrases seen in the document. This strategy disregards the investigation's context.

Machine learning sentiment analysis is based on building models from the training data set. These values are labelled to

927

indicate the degree of orientation of the document. This methodology is often utilized in the opinion collection method for analyzing the pleasing sentiments pertaining to a topic in the data. This strategy can be applied to a variety of things, themes, people, events, and services. This kind of analysis produces exceedingly variable results [18]. These strategies do not appear to be successful due to concerns with word semantic placement, which fluctuates depending on context. Deep learning-based techniques can be utilized efficiently in opinion mining to estimate users, emotions, sentiments, likes, and dislikes [19].

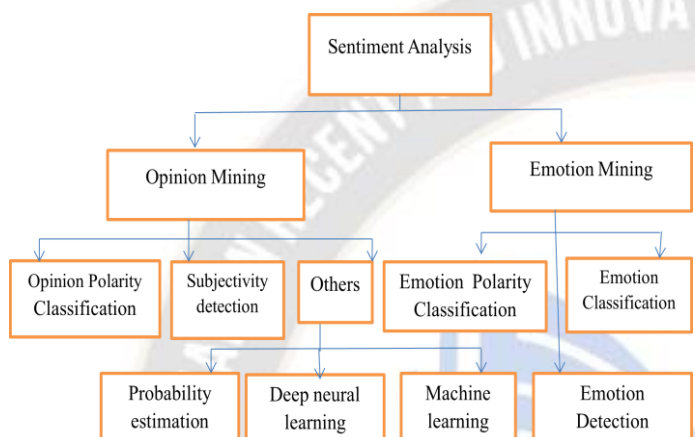

Figure 1. Sentiment analysis of opinion mining.

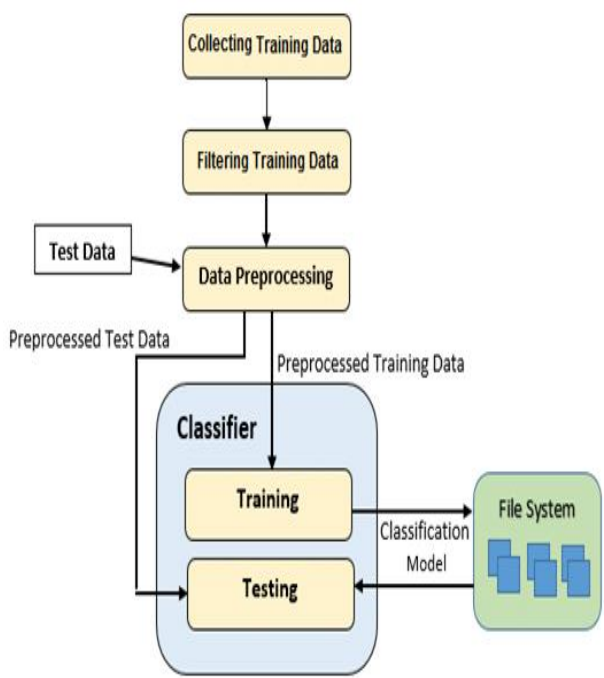

Figure 2. The schematic view of machine learning-based Opinion Mining

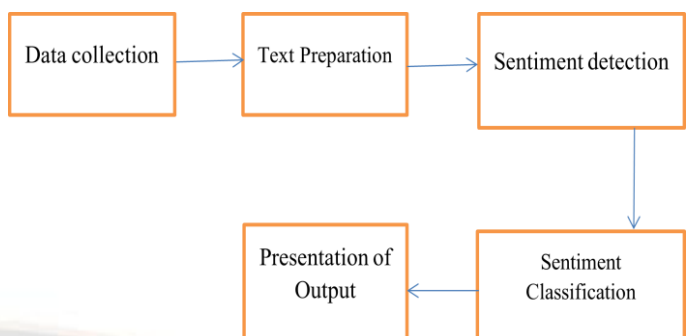

Figure 3. The Process of Opinion Mining

As the amount of unstructured textual data is increasing day by day, the chances and challenges in the field of text mining have drawn the interests of researchers [20]. Several tools are used for exploring them. In this Chapter, Bidirectional Deep Recurrent Neural Networks (BDRNN) are proposed to find the sentiment polarity. A dataset with sentiments is formed for training and testing to extract unbiased opinions. Sentiment Analysis based on Bidirectional Deep Recurrent Neural Networks (SA-BDRNN) Scheme is proposed to deal with the challenges and chances available in text mining.

II. RELATED WORK

The authors, Imran et al. [21], examined the problem of data imbalance within educational feedback datasets and its potential effects on sentiment analysis models. Synthetic text generation models, such as CatGAN and SentiGAN, are employed to address the issue of imbalanced datasets by creating text samples for the underrepresented minority classes. The experimental findings indicate a notable enhancement in the efficacy of sentiment analysis models when applied to imbalanced datasets, following the implementation of synthetic data for balancing purposes. The research additionally examines the assessment metrics employed for assessing the diversity of the generated text and conducts a comparative analysis of CatGAN and SentiGAN in relation to text quality and diversity.

Nayak et al. [22] proposed a modified Bayesian Boosting algorithm with weight-guided optimal feature selection for sentiment analysis. The algorithm effectively addresses the issues of sentiment analysis by utilizing the Bayesian framework and boosting approaches, capitalizing on their inherent strengths. The system integrates a technique for feature selection, which assigns weights to each feature according to their significance in sentiment classification, thereby improving the performance of classification. The approach employs a dynamic adjustment of weights assigned to poor classifiers throughout the boosting iterations, placing greater emphasis on misclassified examples and facilitating more targeted learning for precise sentiment prediction.

In their study, Habbat et al. [23] conducted an analysis of the The duty of sentiment analysis has played a crucial role

**928**

in comprehending client feedback and enhancing commercial tactics. Deep learning models, such as Convolutional Neural Networks (CNNs) and Recurrent Neural Networks (RNNs), have demonstrated encouraging outcomes in the field of sentiment analysis. The researchers have developed a methodology that incorporates GRU and CNN models, in conjunction with XLNet word embeddings, in their study. The combination under investigation exhibits apparent enhancement, as evidenced by experimental findings on three distinct datasets from France. The achieved accuracies on these datasets are 96.5%, 90.1%, and 89.6%.

In their study, Braig et al. [24] undertook a comprehensive investigation of the existing literature to investigate the utilization of machine learning algorithms for sentiment analysis on Twitter data pertaining to the COVID-19 pandemic. The study selected a total of 40 scholarly articles that were published between the time frame of October 2019 to January 2022. These articles employed sentiment analysis as a methodology to assess the prevailing public mood towards various subjects linked to the COVID-19 pandemic. The ensemble models that exhibited the highest accuracy were those that integrated multiple machine learning classifiers, with a particular emphasis on BERT and RoBERTa models that had undergone fine-tuning using Twitter data. The objective of this study is to offer assistance to decision-makers and public health specialists regarding the utilization of sentiment analysis in the management of the pandemic.

In their recent publication, Lee et al. [25] introduced a novel approach to sentiment analysis focused on racism, utilizing deep learning techniques. The study conducted an investigation of racism on social media, encompassing both overt and covert manifestations. This analysis was performed by employing sentiment analysis techniques on a dataset of tweets, utilizing a stacked ensemble GCR-NN model. The GCR-NN model, which integrates the GRU, CNN, and RNN architectures, had exceptional performance, achieving an accuracy of 0.98 and successfully identifying 97% of tweets containing racist remarks. In order to enhance the learning efficiency of machine learning models employed for the identification of racism in tweets, the exclusion of stop words and the removal of noise were implemented. The procedure of tokenization and stemming was employed to preprocess the sample tweets in preparation for analysis.

III. PROPOSED SENTIMENT ANALYSIS TECHNIQUE BASED ON BIDIRECTIONAL DEEP RECURRENT NEURAL NETWORKS (SA-BDRNN)

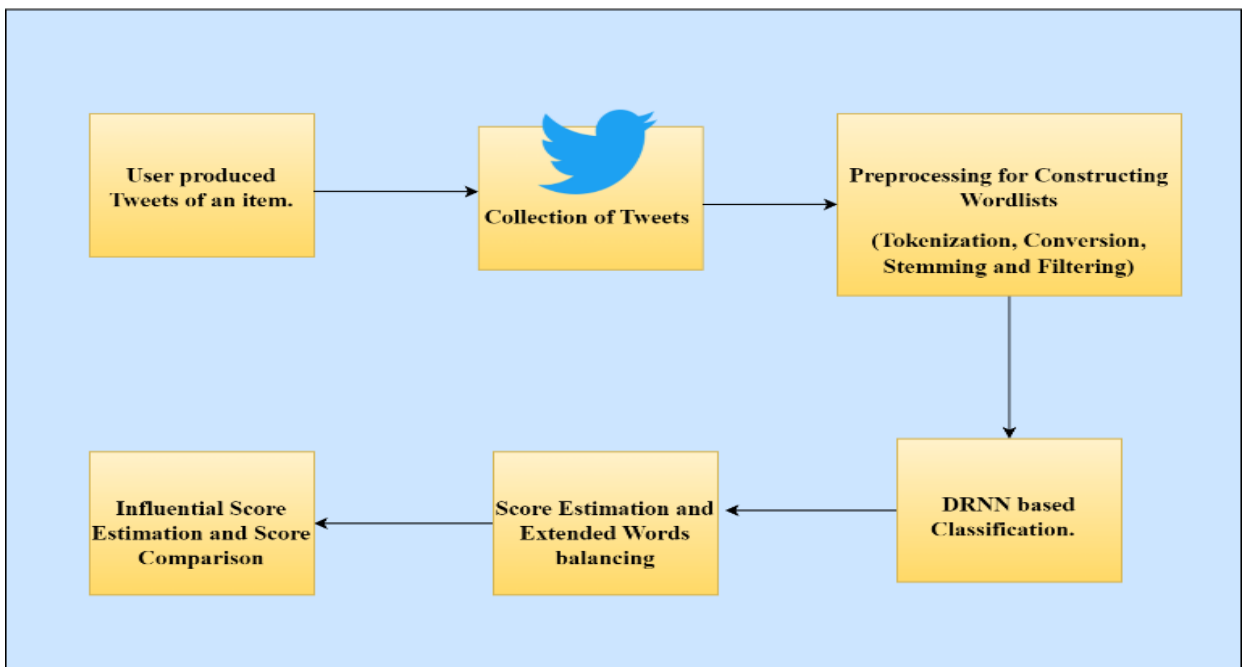

Figure 4. System Model of SA-BDRNN Scheme.

The proposed SA-BDRNN seems to be more trusted machine learning based sentiment analysis scheme in mining opinions. The proposed scheme focuses on dealing with semantic analysis that presents a dynamic approach based on posts on social media and data framework. The behaviors and emotions of users on a topic of interest in real-time are analyzed. Initially, Bidirectional Deep Recurrent Neural Networks (BDRNN) are proposed to find the sentiment polarity. A dataset with sentiments is formed for training and testing to extract unbiased opinions. Sentiment Analysis based on Bidirectional

**929**

Deep Recurrent Neural Networks (SA-BDRNN) Scheme is proposed to deal with the challenges and chances available in text mining. In the proposed SA-BDRNN Scheme, a framework is established to deal with mining opinions for Jallikattu issue in Tamil Nadu. The performance of this proposed SA-BDRNN scheme is compared with the existing frameworks Convolutional Neural Network Deep Learning-based Sentiment Analysis (CNN-DLSA) [26], Aspect Contextual Convolutional Neural Network Deep Learning-based Sentiment Analysis (AC-CNN-SA) [27] and Time Series Analysis-based Deep Neural Network Method (TSA-DNNM) [28,29] schemes used in sentimental analysis.

The proposed SA-BDRNN Scheme includes 4 stages namely,
1. Construction of words representing sentiments
2. Categorizing sentimental words using BDRNN
3. Balancing words preceding implementation of the prediction algorithm.
4. PredictionLet, $X\_i$ $(i = 1 \ldots n)$ Group of items, people or services considered for comparison $T = \{X_1, X_2 \ldots X_n\}$- Focused context.

A.  *Phase 1: Construction of words representing sentiments*

Initially, a repository of words is formed with highly semantically related words based on a particular perspective involving a smaller set of leading hashtags in (Nirmala Devi and Jayanthi 2016). Hashtags are employed.

to classify user posts as positive, negative, or neutral based on the hashtag depiction. Most of the word building methods involve a relatively huge set of hashtags that are manually marked for integrating with word repositories with the goal of improving the classification accuracy. When a greater number of hashtags are taken into consideration, more time will be involved. Hence, the proposed SA-BDRNN Scheme involves a reduced number of hashtags for efficient and probable construction of the word lists.

Words representing sentiments are added to the word lists. The process of repository building involves the following steps.
i. Initially, the posts related to '$X\_i$' are accumulated. The main aim of the proposed SA-BDRNN Scheme emphasizes on finding the most prevalent hashtags based on the degree of happening related to '$X\_i$'. A relatively reduced set of hashtags (2 to 3 in each group) is manually categorized into positive and negative such that the data associated with both the categories are distinctively collected. To be more precise, an upper threshold polarity is set for efficiently categorizing the accumulated posts.
ii. Secondly, the classified hashtags are preprocessed, as the data might contain non-textual data including and misspelt words. Pre-processing involves the following processes, viz., tokenization, conversion, stemming and filtering. In the tokenization phase, tokens like Hypertext Markup Language (HTML) tags, unicode characters, hashtags in Twitter, symbols, phone numbers, Uniform Resource Locator (URL), verbs, nouns, adjectives, adverbs, and emoticons are identified. During the conversion phase, the letters in the words are converted to lowercase and the letters which happen to occur multiple times are modified to be with a single frequency. In morphological stemming, conjugation and plural genders are removed. In filtering, corresponding adjectives and verbs are added to correctly indicate positive and negative sentiments. Positive and negative transitional words representing sentiments are derived.
iii. In this step, positive, negative, and neutral transitional words representing sentiments are refined. This helps in constructing the annotated word repository for '$X\_i$'. Hashtags representing neutral sentiments are removed as they deal with diverse dimensions based on the context. Neutral biasing is applied over the words for the complete set of classes. To be more precise, an empirical test is done to test the values in the range 0.4 and 0.8, to find a threshold that classifies words with an extremely reduced error rate. In the proposed SA-BDRNN Scheme, the threshold is set to 0.7. Hence, the sentiments are set to -1, 0, -1 conforming to negative, neutral, and positive sentiment words.

B.  *Phase 2: Classifying emotional behaviors with BDRNN*

In this proposed SA-BDRNN Scheme, BDRNN s are used for enabling efficient sequential learning that focuses on the task of NLP. Deep Recurrent Neural Networks (DRNN) includes 2 RNNs, in which the first RNN is involved in finding the forward hidden sequences ($F_{hs}^1$), and the second is used for finding the backward hidden sequences ($B_{hs}^2$). The output sequence ($Y_O$) of the DRNN conforming to the input sequence ($X_i$) is obtained using Equation (1).

$$Y_O = A(W_{m(hs)}^1 * F_{hs}^1 + W_{m(hs)}^2 * F_{hs}^2 + O_{BIAS}) \quad (1)$$

Where $A(*)$ - Activation function and $O_{BIAS}$ - Output bias.

Forward and backward activations are shown in the following Equations.

$$F_{hs(n)}^1 = f(W_m^{n-1} * F_{hs(n-1)}^t + W_m^n * F_{hs(n)}^{t-1} + B_{hs(n)}^1) \quad (2)$$

$$F_{hs(n)}^2 = f(W_m^{n-1} * F_{hs(n-1)}^t + W_m^n * F_{hs(n)}^{t+1} + B_{hs(n)}^1) \quad (3)$$

The DRNN involves 3 hidden layers that are adequate to produce improved performance by effective learning of transitional words representing sentiments. Non-conventional language is being used in the social media. The words are

930

purposely transcribed using upper case letters and may include duplications of successive letters termed as extended words. These words are least used in sentiment analysis. Hence in the proposed scheme, the ensuing step deals with balancing and management of extended words.

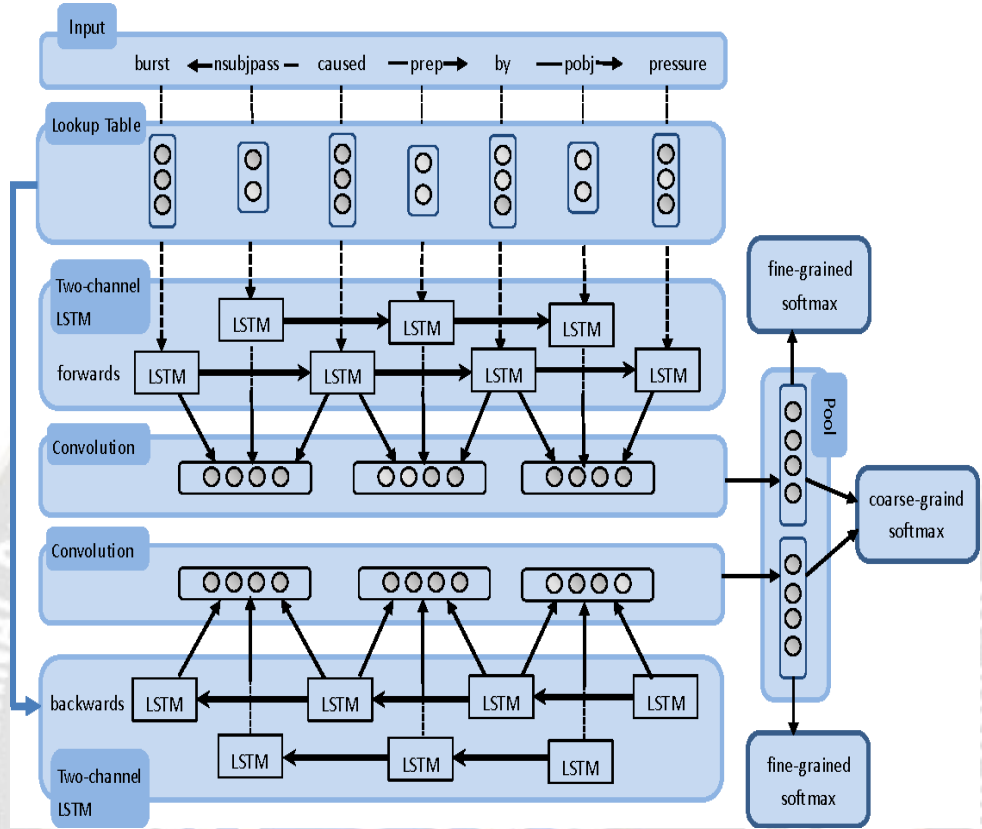

Figure 4. Architecture of BRCNN [31]

### C. Phase 3: Balancing words preceding implementation of the prediction algorithm

The polarity degrees related to opinions are found by incorporating the scores of the associated words at a particular point of time. Messages with specific length are considered as shown in Equation (4)

$$P_t = \sum_{i=1}^{n} WordScore_i \qquad (4)$$

To balance the words before passing them on to the prediction, tweets are categorized into the following:
   i.   Positive - Strong, Moderate, Weak
   ii.  Neutral
   iii. Negative - Strong, Moderate, Weak

An empirical test is carried out to find the threshold of classes.

### D. Phase 4: Prediction

In most of the work existing in the literature, the degree of sentiments was found by finding the positive, negative and neutral sentiments by deriving emotions and words. Thus, the proposed SA-BDRNN scheme considers the difference of class impacts for finding the dominant opinion of the public with the rate related to the corresponding field or subject.

In the context of prediction, each class is allocated a weight ranging from -3 to +3. The given value of "strongly positive" can be represented as +3. The sentiment of the text can be classified as moderately positive, which can be represented by a numerical value of +2. A weakly positive value can be represented as +1. The assigned numerical values for different levels of negativity are as follows: neutral (0), very negative (-3), moderately negative (-2), and weakly negative (-1).

The Degree of Impact (DoI) ascribed for every opinion is found by incorporating metrics that support weight balancing of each opinion as shown in Equation (5).

$$DoI_t = W_t + N_t^L + N_t^{RT} \qquad (5)$$

Where, $W_t$ - Weights related to every class, $N_t^L$ - Number of liked opinions, and $N_t^{RT}$ - Number of Retweeted opinions.

Finally, the total rating associated with the opinions is found based on the collective sum of DoI and the number of tweets exchanged between the consumers about the Item of Interest (IoI) as shown in Equation (6)

$$Rate(X_i) = \frac{\sum_{i=1}^{n} DoI_{t_i}}{N_{PL}} \qquad (6)$$

Where, $N_{PL}$- Number of positive likes. The opinion, that is highly rated, is taken as the maximum likelihood of likes given by consumers over the internet for the item.

## IV. METHODS AND EXPERIENCE

The 'Support for Jallikattu' on social media including Twitter and Facebook are considered to confirm the support and non-support of people discussing on the subject. In this work, the users' views on Twitter are alone considered, wherein users interact and post messages through tweets with a maximum size of 140 characters. REST API is found to suitable as data can be extensively managed by researchers and developers. Permission is granted for collecting available tweets related to a particular hashtag.

### A. Implementation

The suggested system involves the implementation of a cluster consisting of three servers. This is done to meet the requirements of sentiment-based data analysis, which necessitates sufficient memory capacity and the ability to execute parallel processing. The servers are equipped with two Intel Xeon E5530 Quad core CPU 2.4 GHz processors, which have the capability to execute within a 64-bit Linux Ubuntu Environment. The servers are equipped with a 1 terabyte hard disc drive and 24 gigabytes of DDR3 random access memory. Apache Kafka is utilized for data collection due to its extensive streaming platform capabilities, which encompass functions such as message publication, subscription, and forwarding. These functionalities ensure the reliable and distributed delivery of messages, hence enabling a continuous and scalable service. To achieve precision, a cohesive collection of API libraries is utilized for the purpose of constructing applications that are well-suited for efficient stream management. Moreover, a substantial amount of data is stored within the Hadoop Distributed File System (HDFS).

The Spark is used in data processing that supports the building of word lists and classification of data. Spark supports scalable and fault tolerant handing of streams in succeeding time slots. The Spark is used with Hadoop and managed using YARN over 3 servers. DRNN is implemented using the benefits of Spark MLlibs as the performance of the proposed and the existing approaches are to be compared. The algorithms that support data collection and processing are executed through Python. Tweets are retrieved in real-time using DRNN that is modeled using Spark MLlibs.

### B. Data Collection

Data collection involves the following phases.
**Phase 1:**
Keywords like 'Support' and 'Non-support' are employed in reclaiming tweets that are classified into.

$$P = \{a_1, a_2\} \qquad (7)$$

where, $a_1$ – Support and $a_2$ - Non-Support.

Data is collected using Twitter RESTAPI in real-time with tweets in English. Appropriate hashtags are gathered based on the word frequency list so as to measure the maximum prevalent hashtags for the collected data. Tweets are classified as discussed below

- Positive hashtags to support Jallikatu: #tamizhan history, # tamizhan gethu, #tamizhan identity, # tamizhan super.
- Hashtags to oppose Jallikatu: #jallikatu not safe, #lifesucking jallikatu, #risky jallikatu, # dangerous jallikatu.

To validate the process of word list construction, data related to the above-mentioned hashtags are collected during the period December 21 & 22, 2016. The data includes 130,000 tweets segregated into 2 categories include positive and negative support with 65,000 tweets each.

**Phase 2:**
A prediction is made to assess the level of support for Jallikattu by collecting messages from Twitter on December 21 and 22, 2016. Around 32,00,000 tweets were collected for both support and opposition to the subject to analyze it.

**Phase 3:**
The degree of polarity connected to the Tweets is computed in order to determine the Degree of Impact (DoI) and Weight.

**Phase 4:**
This stage takes the data from the previous stage and replaces it with tweets that include the Class Weight and Labelled class for which the Tweet corresponds to the number of likes and retweets.

### C. Data Processing

In the initial phase, word lists comprising positive and negative sentiments are constructed to assess the levels of support and non-support towards the practice of Jallikatu. The data undergoes pre-processing through the application of filters, which involves the removal of duplicated Tweets and stop words. The process of classification is performed utilizing a Deep Recurrent Neural Network (DRNN), wherein the inclusion of additional terms beyond the core vocabulary is considered. Tweets are categorized through the process of training. The polarity degree is determined, afterwards followed by the Degree of Impact (DoI). The comprehensive rating is calculated to determine the level of support and non-support for the subject of investigation, as illustrated in the subsequent tables.

TABLE 1. TWEET SAMPLES TAKEN FOR ANALYSIS

| Positive Tweet Hashtags Queried | Negative Tweet Hashtags Queried |
|---|---|
|  |  |

**932**

| #tamizhan history, #tamizhan identity, # tamizhan super, # tamizhan gethu | #jallikatu not safe, #lifesucking jallikatu, #risky jallikatu, # dangerous jallikatu |
|---|---|

TABLE 2. PRE-PROCESSING AND CLASSIFICATION OF TWEETS USING DRNN

| Classifier used for learning tweets | Output | Action incorporated |
|---|---|---|
| Preprocessing | #tamizhan history, #tamizhan identity, # tamizhan super, # tamizhan gethu | Extended words are handled |
| Deep Recurrent Neural network | Balancing and handling of extended words through the process of training | Learning the intermediate sentiment words for estimating degree of polarity |

TABLE 3. POLARITY ESTIMATION BY BALANCING THE COLLECTION OF WORDS

| Processes | Value | Output |
|---|---|---|
| Degree of polarity | 3 | Strongly positive |

TABLE 4. PREDICTION

| Processes | Value | Output |
|---|---|---|
| Process 1 | Degree of Influence | 4 |
| Process 2 | Overall rating | 38 |

## V. SIMULATION RESULTS AND DISCUSSIONS

The performance of the proposed SA-BDRNN Scheme is evaluated by conducting simulation experiments and analysis for different parameters including Size of Vector [30] and Vocabulary, number of hidden layers [31], Count and Size of filters, drop out, Regularizer and Activation functions as shown in table 5.

TABLE 5. TWEET SAMPLES TAKEN FOR ANALYSIS

| Parameters | Value |
|---|---|
| Input Vector Size | 100 |
| Vocabulary Size | 13, 398 |
| Filter Size | 3, 4, 7 |
| Number of filters | 10,40,70,100,28,256 |
| Output Dimensions | 128 |
| Regularizer value | 1.3 |
| Drop out value | 0.4 |
| Batch Size | 64 |
| Count of Epochs | 5 |
| Number of Hidden Layers | 4,5 |
| Number of Recurrent Layers | 2,3 |

Initially, the prevalence of the proposed SA-BDRNN Scheme is established for conducting sentimental analysis. The performance is appraised based on the Accuracy and Loss curve for varying number of training iterations. Figure 5 shows the comparison of accuracies of the proposed SA-BDRNN Scheme and the existing [26], [27] and [28] for varying number of training iterations. It is observed that the accuracy of the proposed scheme is persistent over the benchmarked schemes used for Sentiment analysis due to classification using DRNN.

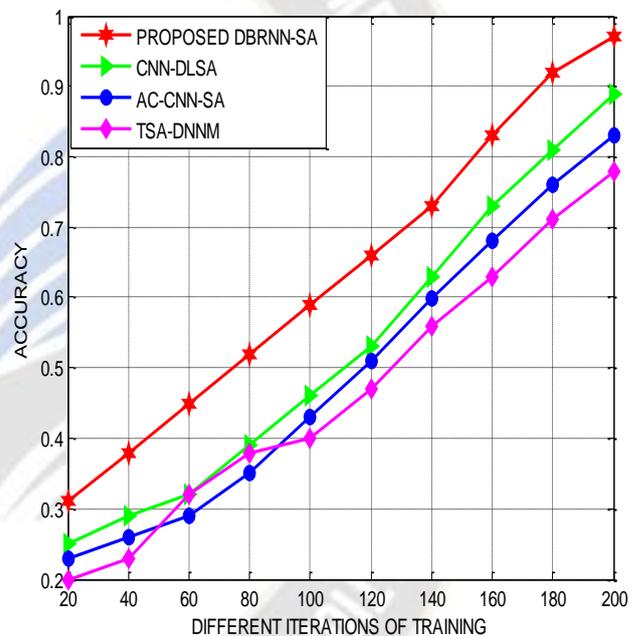

Figure 5 The Proposed SA-BDRNN Scheme's Accuracy for Varying Training Iterations.

The proposed SA-BDRNN Scheme offers 12%, 10% and 7% better accuracies in contrast to [26], [27] and [28] schemes respectively. Similarly, loss curve of the proposed SA-BDRNN Scheme is compared with that of the existing [26], [27] and [28] for varying number of training iterations. The proposed SA-BDRNN scheme offers reduced loss of errors in contrast to the benchmarked schemes used for sentiment analysis. This is made possible by inheritance in the stemming process during preprocessing. The SA-BDRNN Scheme, as depicted in Figure 6, demonstrates a reduction in loss error of 14%, 11%, and 8% compared to the [26], [27] and [28] schemes, respectively.

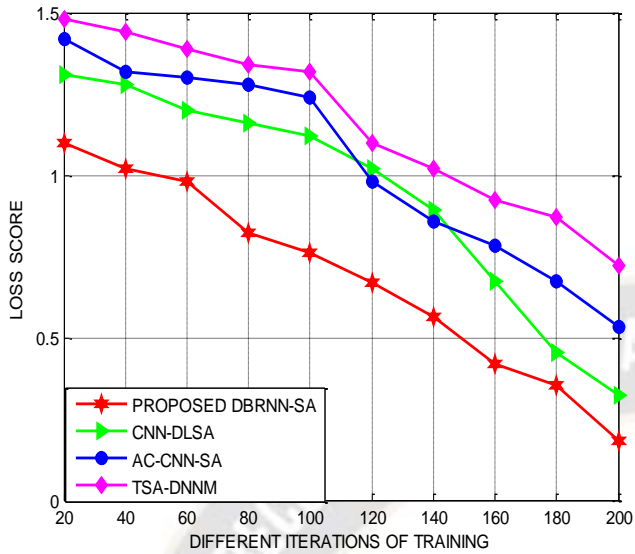

Figure 6 The Proposed SA-BDRNN Scheme's Loss Curve for Varying Training Iterations.

The second phase of the study involves quantifying the prevalence of the proposed SA-BDRNN Scheme by measuring Precision and Recall across different numbers of training iterations. Figure 7 illustrates the Precision of the SA-BDRNN Scheme, as compared to the [26], [27] and [28] schemes, across different numbers of training iterations. The stability of the suggested SA-BDRNN Scheme is seen because to its incorporation of pre-processing, which enhances precision. The SA-BDRNN Scheme demonstrates a significant improvement in precision compared to the [26], [27] and [28] schemes, with respective increases of 12%, 10%, and 7%.

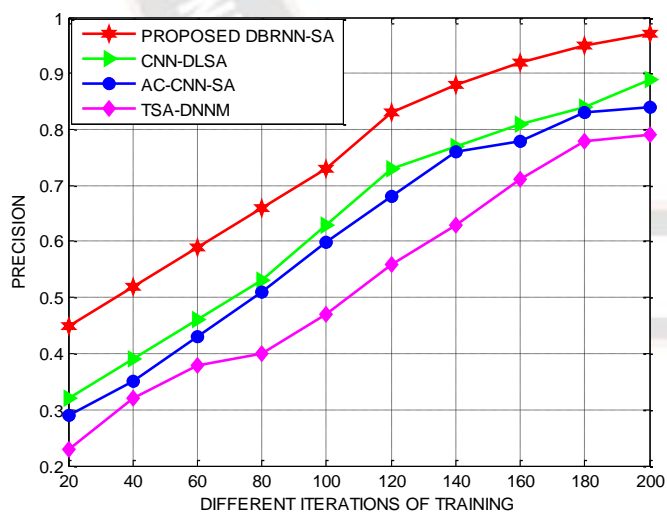

Figure 7. Precision of the Proposed SA-BDRNN Scheme for Varying Training Iterations.

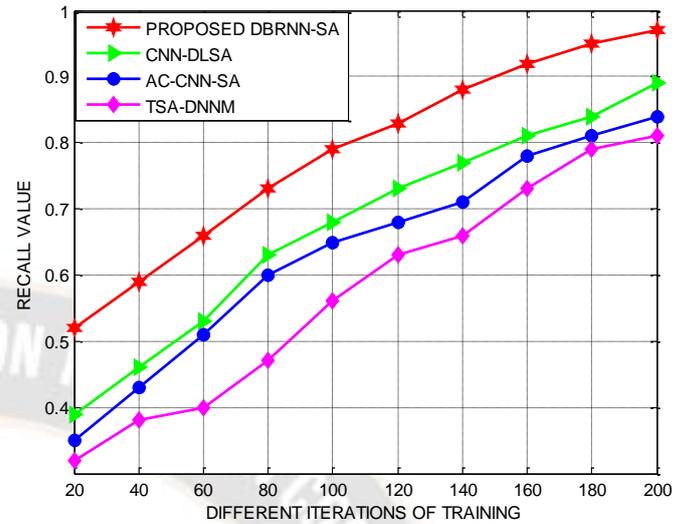

Figure 8. The Proposed SA-BDRNN Scheme's Recall Value for Varying Training Iterations.

Figure 8 illustrates the Recall performance of the SA-BDRNN Scheme, as well as the existing [26], [27] and [28] schemes, over different numbers of training rounds. The proposed technique demonstrates improved Recall value by effectively identifying and incorporating both dependent and independent components that contribute to sentiment analysis to their fullest extent. The SA-BDRNN Scheme exhibits a reduction of 13%, 11%, and 9% in Recall values compared to the [26], [27] and [28] schemes, respectively.

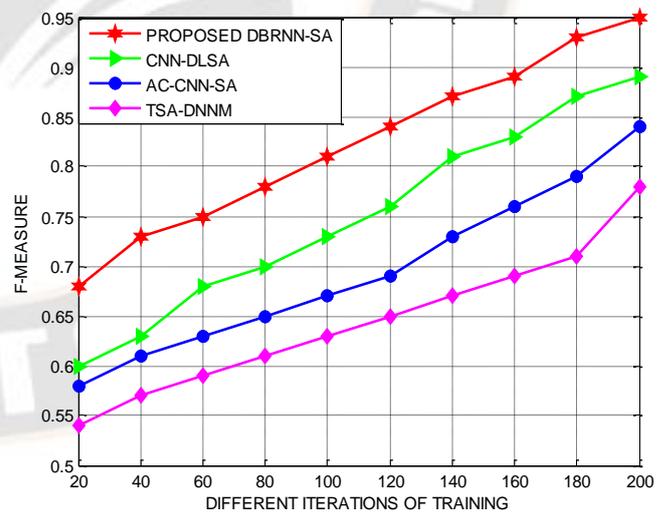

Figure 9. The proposed SA-BDRNN Scheme's F-Measure for Varying Training Iterations.

In the concluding phase of the inquiry, the superiority of the proposed SA-BDRNN approach is demonstrated by evaluating its performance using F-Measure and Mean Absolute Error (MAE) metrics across different numbers of training iterations. Figure 9 illustrates the F-Measure values obtained from the

**934**

proposed SA-BDRNN Scheme, as well as the existing [26], [27] and [28] schemes. The SA-BDRNN Scheme is seen to exhibit independence from the number of training iterations due to the implementation of word balancing prior to inputting them into the prediction algorithm as shown in Figure 9. The SA-BDRNN Scheme demonstrates improved performance in terms of F-Measure compared to previous schemes such as [26], [27] and [28] with enhancements of 13%, 11%, and 8% respectively.

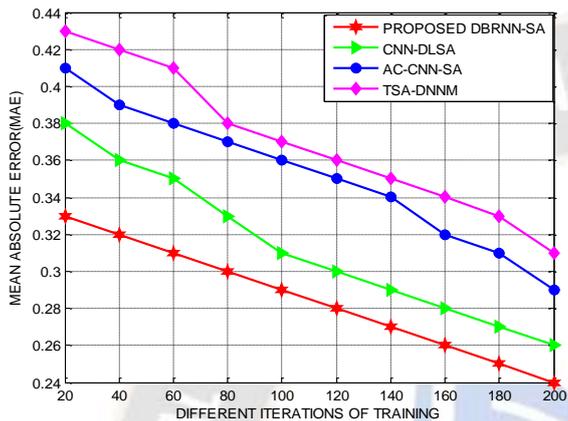

Figure 10. The Mean Absolute Error (MAE) of the Suggested SA-BDRNN Protocol for Intermittent Training.

Figure 10 illustrates the Mean Absolute Error (MAE) of the SA-BDRNN Scheme, as well as the benchmark schemes [26], [27] and [28] across different numbers of training iterations. The proposed scheme of SA-BDRNN exhibits a decrease in the Mean Absolute Error (MAE) because of the thorough examination of several well-known linguistic aspects. The SA-BDRNN Scheme, when compared to the [26], [27] and [28] schemes, demonstrates a reduction in mean absolute error (MAE) of 11%, 8%, and 6% respectively.

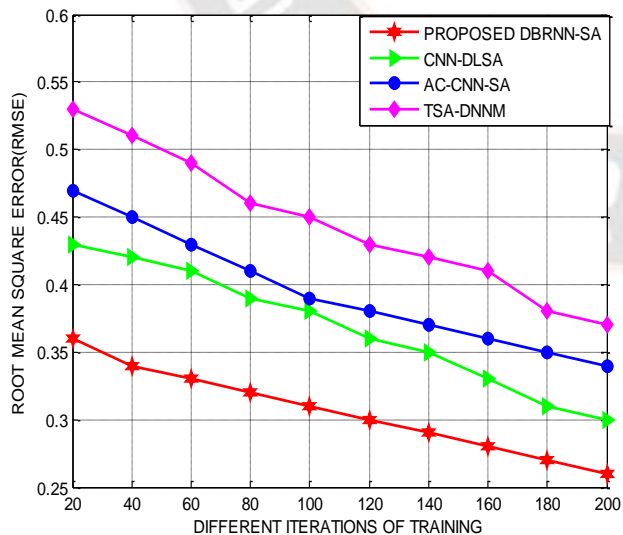

Figure 11. RMSE of the Proposed SA-BDRNN Scheme for Varying Training Iterations.

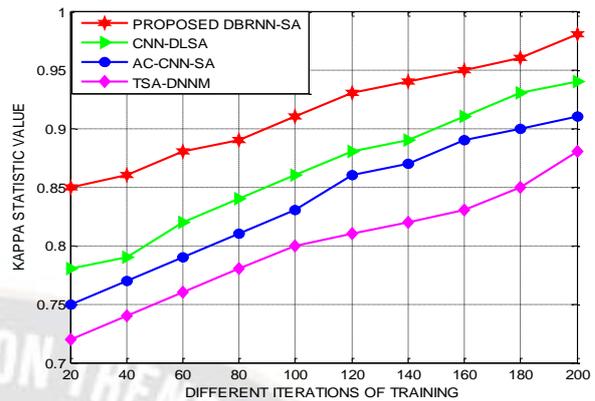

Figure 12. Kappa Statistic-the Proposed SA-BDRNN Scheme for Varying Training Iterations.

The significance of the suggested SA-BDRNN Scheme is determined by assessing its performance using evaluation metrics such as Root Mean Square Error (RMSE), Kappa Statistics, and True Positive rate over different training iterations. Figure 11 illustrates the root mean square error (RMSE) values of the SA-BDRNN Scheme, as well as the [26], [27] and [28] schemes that are currently in existence. The proposed scheme demonstrates a decrease in root mean square error (RMSE) for different numbers of training iterations, as they have the ability to inherit. The SA-BDRNN Scheme exhibits a reduction in RMSE of 15%, 13%, and 10% compared to the currently used schemes, respectively.

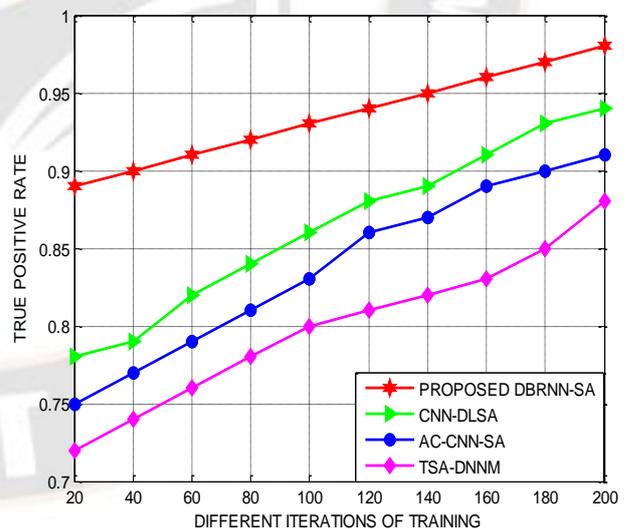

Figure 13. True Positive Rate-Proposed SA-BDRNN Scheme for Varying Training Iterations.

The Kappa statistics for different numbers of training iterations are depicted in Figure 12. The SA-BDRNN Scheme, if implemented, demonstrates improved performance in terms of Kappa value. This is attributed to the utilization of precise classification methods that effectively distinguish the correct

**935**

polarity of feelings derived from tweets. The SA-BDRNN Scheme demonstrates a significant improvement in Kappa Statistic compared [26], [27] and [28] to the current schemes, with enhancements of 14%, 11%, and 10% respectively. The Figure 13 shows the true positive comparison against the prevailing schemes and proved that the proposed scheme outperforms all in with different iteration of training.

## VI. CONCLUSION AND FUTURE WORK

In this work, SA-BDRNN Scheme incorporating Deep Recurrent Neural Network based learning is proposed to envisage user's behavior by analyzing social media data. In the proposed scheme, a word list is built using the polarity for a limited set of hash tags related to a topic. The sentiments are classified using DRNN and the extended words are balanced for efficient classification. From the results, it is evident that the proposed SA-BDRNN scheme offers 11% and 13% better Kappa statistics and True positive rate for varying number of training iterations. In the future, a unified model can be built with Convolutional Neural Network (CNN) and Long Short-Term Memory (LSTM) for analyzing sentiments involving numerous linguistics.